\pdfoutput=1

\documentclass[11pt]{article}

\usepackage[backref=page]{hyperref}

\usepackage{acl}

\usepackage{times}
\usepackage{latexsym}

\usepackage[T1]{fontenc}

\usepackage[utf8]{inputenc}

\usepackage{graphicx}
\usepackage{multirow}
\usepackage{enumitem}
\usepackage{booktabs}
\usepackage{listings}
\usepackage[nameinlink,capitalize,noabbrev]{cleveref}

\usepackage{microtype}

\crefformat{footnote}{#2\footnotemark[#1]#3}

\title{EventGraph at CASE 2021 Task 1: A General Graph-based Approach to Protest Event Extraction}

\author{Huiling You,$^1$ David Samuel,$^1$ Samia Touileb,$^2$ \and Lilja Øvrelid$^1$ \\
         $^1$University of Oslo\\
         $^2$University of Bergen \\ 
         \texttt{\{huiliny, davisamu, liljao\}@ifi.uio.no} \\
         \texttt{samia.touileb@uib.no}
}

\begin{document}
\maketitle
\begin{abstract}
This paper presents our submission to the 2022 edition of the CASE 2021 shared task 1, subtask 4. The EventGraph system adapts an end-to-end, graph-based semantic parser to the task of Protest Event Extraction and more specifically subtask 4 on event trigger and argument extraction. We experiment with various graphs, encoding the events as either ``labeled-edge'' or ``node-centric'' graphs. We show that the ``node-centric'' approach yields best results overall, performing well across the three languages of the task, namely English, Spanish, and Portuguese. EventGraph is ranked 3rd for English and Portuguese, and 4th for Spanish. Our code is available at: \url{https://github.com/huiling-y/eventgraph_at_case}

\end{abstract}

\section{Introduction}
The automated extraction of socio-political event information from text constitutes an important NLP task, with a number of application areas for social scientists, policy makers, etc.
The task involves analysis at different levels of granularity: document-level, sentence-level, and the fine-grained extraction of event triggers and arguments within a sentence.
The CASE 2022 Shared Task 1 on Multilingual Protest Event Detection extends the 2021 shared task \cite{hurriyetoglu-etal-2021-multilingual} with additional data in the evaluation phase and features four subtasks: (i) document classification, (ii) sentence classification, (iii) event sentence co-reference, and (iv) event extraction. 

The task of event extraction involves the detection of explicit event triggers and corresponding arguments in text. Current classification-based approaches to the task typically model the task as a pipeline of classifiers \citep{ji-grishman-2008-refining, li-etal-2013-joint, liu-etal-2020-event, du-cardie-2020-event, li-etal-2020-event} or using joint modeling approaches \citep{yang-mitchell-2016-joint, nguyen-etal-2016-joint-event, liu-etal-2018-jointly, wadden-etal-2019-entity, lin-etal-2020-joint}.

In this paper, we present the EventGraph system and its application to Task 1 Subtask 4 in the 2022 edition of the CASE 2021 shared task.
EventGraph is a joint framework for event extraction, which encodes events as graphs and solves event extraction as semantic graph parsing.
We show that it is beneficial to model the relation between event triggers and arguments and approach event extraction via structured prediction instead of sequence labelling. Our system performs well on the three languages, achieving competitive results and consistently ranked among the top four systems. 

In the following, we briefly describe the data supplied by the shared task organizers and present Subtask 4 in some more detail. We then go on to present an overview of the EventGraph system focusing on the encoding of the data to semantic graphs and the model architecture. We experiment with several different graph encodings and provide a more detailed analysis of the results.

\section{Data and task}
Our contribution is to subtask 4, which falls under shared task 1 -- the detection and extraction of socio-political and crisis events. While most subtasks of shared task 1 have sentence-level annotations, subtask 4 has been annotated at the token-level while providing the annotators the document-level contexts. Subtask 4 focuses on the extraction of event triggers and event arguments related to contentious politics and riots \cite{hurriyetoglu-etal-2021-multilingual}. This subtask has been previously approached as a sequence labeling problem combining various methods of fine-tuning pre-trained language models \cite{hurriyetoglu-etal-2021-multilingual}.

\begin{figure*}[t!]
\centering
\includegraphics[width=\textwidth]{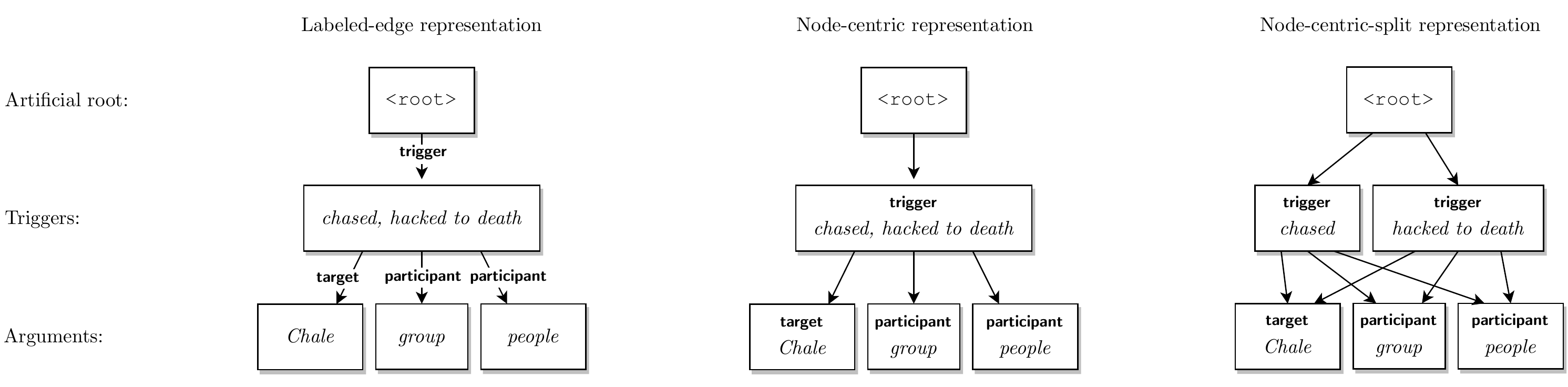}
\caption{Graph representations of sentence ``\textit{Chale was allegedly chased by a group of about 30 people and was hacked to death with pangas, axes and spears.}'' }
\label{fig:graphs}
\end{figure*}

The data supplied for Subtask 4 is identical to that of the 2021 edition of the task, as presented in \citet{hurriyetoglu-etal-2021-multilingual}. The data is part of the multilingual extension of the GLOCON dataset \cite{hurriyetouglu2021cross} with data from English, Portuguese, and Spanish. The source of the data is protest event coverage in news articles from specific countries: China and South Africa (English), Brazil (Portuguese), and Argentina (Spanish). 
The data has been doubly annotated by graduate students in political science with token-level information regarding event triggers and arguments. \citet{hurriyetoglu-etal-2021-multilingual} reports the token level inter-annotator agreement to be between 0.35 and 0.60. Disagreements between annotators were subsequently resolved by an annotation supervisor.
\cref{tab:data-events} shows the number of news articles for each of the languages in the task, distributed over the training and test sets. This clearly shows that the majority of the data is in English with only a fraction of articles in Portuguese and Spanish.


Relevant statistics for the different event component annotations for Subtask 4 are presented in \cref{tab:data-events} detailing the number of triggers, participants, and various other types of argument components, such as place, target, organizer, etc. Once again, the table also illustrates the comparative imbalance in data across the three languages.
\begin{table}[t!]
\resizebox{\columnwidth}{!}{
\begin{tabular}{@{}lrrr@{}}
\toprule
& {\bf English} & {\bf Portuguese} & {\bf Spanish} \\
\midrule
train & 732 (2,925) & 29 (78) & 29 (91) \\
dev & 76 (323) & 4 (9) & 1 (15) \\
test & 179 (311) & 50 (190) & 50 (192) \\ \midrule
trigger & 4,595 & 122 & 157 \\ 
participant & 2,663 & 73 & 88 \\
place &  1,570 & 61 & 15\\
target & 1,470 & 32 & 64 \\
organizer & 1,261 & 19 & 25 \\
etime & 1,209 & 41 & 40\\
fname & 1,201 & 48 & 49\\
\bottomrule
\end{tabular} %
}
\caption{\label{tab:data-events} \textbf{Top}: Number of articles (sentences) for the different languages in Subtask 4 \cite{hurriyetoglu-etal-2021-multilingual}. About 10 percent (in terms of sentences) of the official training data is used as the development split. \textbf{Bottom}: Counts for the different event components in Subtask 4 training data for English, Portuguese, and Spanish \cite{hurriyetoglu-etal-2021-multilingual}.}
\end{table}

\section{System overview}

We use our system, EventGraph, that adapts an end-to-end graph-based semantic parser to solve the task of extracting socio-political events. In what follows, we give more details about the graph representation and the model architecture of our system. 

\subsection{Graph representations}

We represent each sentence as an event graph, which contains event trigger(s) and arguments as nodes. In an event graph, edges are constrained between the trigger(s) and the corresponding arguments. However, since our system can take as input graphs in a general sense the precise graph representation that works best for this task must be determined empirically. We here explore two different graph encoding methods, where the labels for triggers and arguments are represented either as edge labels or node labels, namely ``labeled-edge'' and ``node-centric''. Since sentences in the data may contain information about several events with arguments shared across these, we also experiment with a version of the ``node-centric'' approach where multiple triggers give rise to separate nodes in the graph. 
The intuition behind this is that it is easier for the model to predict a node anchoring to a single span than to several disjoint spans.

\begin{itemize}\itemsep0em 
    \item \textbf{Labeled-edge}: labels for event trigger(s) and arguments are represented as edge labels; multiple triggers are merged into one node, as shown by the first graph of \cref{fig:graphs}.
    \item \textbf{Node-centric}: labels for event trigger(s) and arguments are represented as node labels; there is always a single node for trigger(s), as shown by the second graph of \cref{fig:graphs}.
    \item \textbf{Node-centric-split}: node labels denote trigger(s) and argument roles; multiple triggers are represented in different nodes, as shown by the third graph of \cref{fig:graphs}.
\end{itemize}

\subsection{Model architecture}

\begin{figure}[t!]
\centering
\includegraphics[width=\columnwidth]{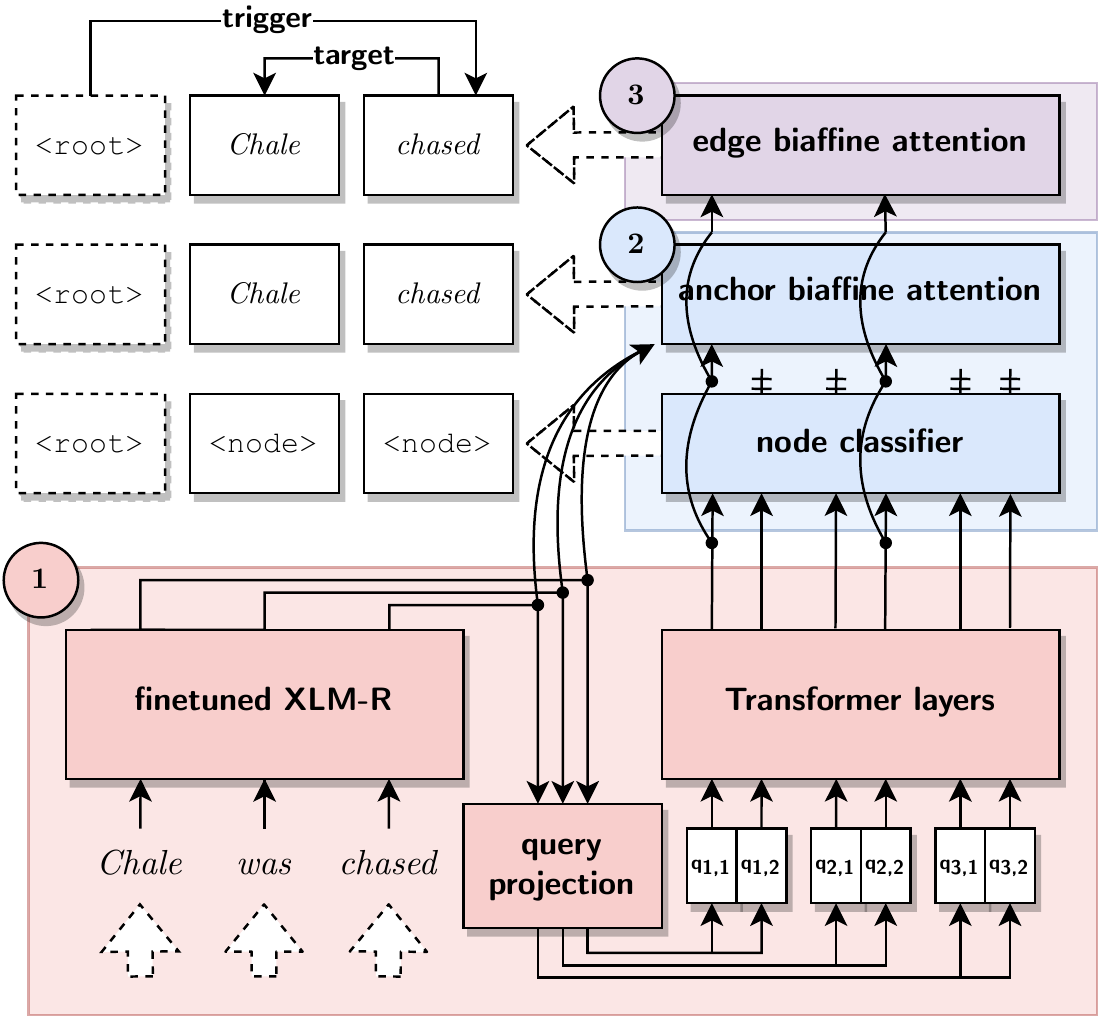}
\caption{\label{fig:perin} \footnotesize EventGraph architecture. 1) the input gets a contextualized representation from \textbf{the sentence encoding} module, 2) graph nodes are decoded by \textbf{the node prediction} module and 3) connected by \textbf{the edge prediction} module. The given example is for ``label-edge'' event graph parsing.}
\end{figure}

Our model is built upon a winning framework \citep{samuel-straka-2020-ufal} from a previous meaning representation parsing shared task  \citep{oepen-etal-2020-mrp}. The model contains customizable components for predicting nodes and edges, thus generating event graphs for different graph representations. We introduce each component of the model as following (\cref{fig:perin}):

\paragraph{Sentence encoding} Each token of an input sentence obtains a contextualized embedding from a pretrained language model, the large version of XLM-R \citep{conneau-etal-2020-unsupervised} in our implementation. These embeddings are mapped onto latent queries by a linear transformation layer, and processed by a stack of Transformer layers \citep{vaswani2017attention} to model the dependencies between queries. 

\paragraph{Node prediction} A node-presence classifier processes the queries and predicts nodes by classifying each query. An anchor biaffine classifier \citep{dozat2017deep} creates anchors from the nodes to surface strings via deep biaffine attention between the queries and the contextual embeddings.

\paragraph{Edge prediction} With predicted nodes, two biaffine classifiers are used to construct the edges between nodes: one classifier predicts the presence of edge between a pair of nodes and the other predicts the corresponding edge label.

The graph generated for each input sentence contains the extracted event components. We then convert the labels to BIO format.

\section{Experimental setup}

\paragraph{Data} We use all the official training data to train our final model, without using any additional data. During development time, we set aside about 10 percent of the training data for development. A breakdown of the number of articles and sentences in train and dev are provided in \cref{tab:data-events}.

\paragraph{Joint training} We train our model on the training data of all three languages and test on the official test data. As shown in \cref{tab:data-events}, the training data for Portuguese and Spanish makes only a small portion of all training data, which leads to few-shot learning for these two languages.

\paragraph{Implementation details} We use the large version of XLM\mbox{-}R via HuggingFace \texttt{transformers} library \citep{wolf-etal-2020-transformers}. All models were trained with a single Nvidia RTX3090 GPU.

\paragraph{Evaluation metrics} The evaluation metric is a macro $F_1$ score for individual languages. The predicted event-annotated texts are in BIO format, and the scores are calculated with a python implementation\footnote{\url{https://github.com/sighsmile/conlleval}} of the \texttt{conlleval} evaluation script used in CoNLL-2000 Shared Task \citep{tjong-kim-sang-buchholz-2000-introduction}, where precision, recall and $F_1$ scores are calculated for predicted spans against the gold spans and there is no dependency between event arguments and triggers.

\begin{table*}[!th]
\centering
\resizebox{\textwidth}{!}{
\begin{tabular}{@{}l@{\hspace{2em}}l@{\hspace{3em}}rrrrrrr@{\hspace{4em}}r@{}}
\toprule 
\textbf{Language} & \textbf{System} & \textbf{trigger} & \textbf{target} & \textbf{Place} & \textbf{Participant} & \textbf{Organizer} & \textbf{fname} & \textbf{etime} & \textbf{all} \\
\midrule

\multirow{4}{*}{En} & & \emph{457} & \emph{134} & \emph{118} & \emph{293} & \emph{131} & \emph{129} & \emph{121} & \\
& Label-edge & 82.48 & 56.29 & 75.44 & 74.62 & 74.52 & 50.42 & 77.06 & 73.46  \\
& Node-centric & 84.21 & 62.09 & 74.89 & 76.42 & 75.46 & 54.31 & 81.22  & 75.85 \\
& Node-centric-split & 84.62 & 52.88 & 75.11 & 73.75 & 74.91 & 52.28 & 78.97 & 73.92 \\ \midrule

\multirow{4}{*}{Es} & & \emph{28} & \emph{5} & \emph{5} & \emph{7} & \emph{4} & \emph{7} & \emph{5} & \\
& Label-edge & 66.67 & 60.00 & 100.00 & 100.00 & 66.67 & 71.43 & 80.00  & 73.85 \\
& Node-centric & 65.62 & 72.73 & 100.00 & 100.00 & 80.00 & 76.92 & 80.00 & 75.76 \\
& Node-centric-split & 71.19 & 54.55 & 100.00 & 100.00 & 66.67 & 85.71 & 60 & 75.59\\ \midrule

\multirow{4}{*}{Pr} & & \emph{11} & \emph{7} & \emph{3} & \emph{5} & \emph{2} & \emph{2} & \emph{5} & \\
& Labeled-edge & 83.33 & 71.43 & 75.00 & 90.91 & 66.67 & 100.00 & 66.67  & 78.87\\
& Node-centric & 88.00  & 61.54 & 66.67 & 90.91 & 100.00 & 100.00 & 66.67 & 79.45\\
& Node-centric-split & 91.67 & 71.43 & 50 & 90.91 & 100.00 & 66.67 & 100.00 & 83.78 \\
\bottomrule

\end{tabular}
}
\caption{\label{tab:results-dev} Detailed $F_1$ scores of our systems on the development data with different graph representations. We also add the number of each event component to better compare the distribution of components against the scores.}
\end{table*}


\paragraph{Submitted systems} We submitted three models as listed in \cref{tab:results}.

\section{Results and discussion}

\begin{table}[t!]
\centering
\resizebox{\columnwidth}{!}{%
\begin{tabular}{@{}l@{\hspace{2em}}l@{\hspace{3em}}r@{}}
\toprule 
\textbf{System} & \textbf{Language} & \textbf{Macro $F_1$} \\
\midrule

\multirow{3}{*}{Labeled-edge} & English & 73.12\hphantom{\textsubscript{3}}   \\
& Spanish & 64.02\hphantom{\textsubscript{3}}    \\
& Portuguese & 69.62\hphantom{\textsubscript{3}}   \\ \midrule
\multirow{3}{*}{Node-centric} & English & 74.02\hphantom{\textsubscript{3}}  \\
& Spanish & 64.16\hphantom{\textsubscript{3}}  \\
& Portuguese & 70.73\hphantom{\textsubscript{3}}   \\ \midrule
\multirow{3}{*}{Node-centric-split} & English &74.76\textsubscript{3} \\
& Spanish &  64.49\textsubscript{4} \\
& Portuguese &  71.72\textsubscript{3} \\ \midrule
\multirow{3}{*}{Winning systems} & English & 77.46\textsubscript{1} \\
& Spanish &  69.87\textsubscript{1} \\
& Portuguese & 74.57\textsubscript{1} \\
\bottomrule

\end{tabular}%
}
\caption{\label{tab:results} Results of our systems on the official test data with different graph representations. We also include the winning system results from the shared task leaderboard. Subscripts indicate the ranking on the leaderboard, so we only add corresponding ranking to our best-performing system.}
\end{table}



\begin{table}[t]
\resizebox{\columnwidth}{!}{%

    \centering
    \begin{tabular}{llrrr} 
    \toprule
    \textbf{Argument}   & \textbf{System}  &  \textbf{P} & \textbf{R} & \textbf{$F1$} \\ \midrule
    \multirow{3}{*}{fname} & Labeled-edge   & 47.62 & 53.57 & 50.42  \\
     & Node-centric & 52.50 & 56.25 & 54.31 \\
    & Node-centric-split & 48.84 & 56.25 & 52.28 \\ \midrule
    \multirow{3}{*}{target} & Labeled-edge & 60.28 & 52.80 & 56.29  \\ 
    & Node-centric & 65.52 & 59.01 & 62.09  \\
    & Node-centric-split & 58.21 & 48.45 & 52.88  \\
    
    \bottomrule
    \end{tabular}
    }
    \caption{Detailed Precision, Recall, and $F1$ scores of \texttt{fname} and \texttt{target} arguments for English developmentset.}
    \label{tab:res_detailed}
\end{table}

We summarize the results of our systems on the official test data in \cref{tab:results}. All scores are obtained by submitting our test predictions to the shared task.\footnote{\label{footnote:codalab}\url{https://codalab.lisn.upsaclay.fr/competitions/7126}, accessed on September 29, 2022.}  Results show that ``node-centric'' systems generate better results than ``label-edge'' systems, and it is more beneficial to keep multiple event triggers as separate nodes. In terms of languages, all models perform best on English, which is unsurprising, since the training data consists mostly of English. However, the results on Portuguese are consistently better than those of Spanish, signaling English might be a better transfer language for Portuguese than for Spanish.

Compared with other participating systems, in particular the winning systems,\cref{footnote:codalab} 
as shown in \cref{tab:results}, our results are still competitive. We rank 3rd for English and Portuguese, and 4th for Spanish; our best results are achieved by a single system. For English and Portuguese, our results are very close to the winning results, which are achieved by different participating systems.

\subsection{Error analysis on development data}

Since the gold data for the test set is not available to task participants, we are not able to perform more detailed error analysis. Hence, to have more insights into our models' performance, we provide some error analysis on the development data (as described in \cref{tab:data-events}). As previously mentioned, during our model development phase, we did not use all the official training data for training, but set aside small set for validation (about 10\%). 

As shown in \cref{tab:results-dev}, over all event components, \texttt{target} and \texttt{fname} arguments are more difficult to extract than others, with the scores substantially lower across different languages and models. In general, our models perform best in \texttt{trigger} extraction, partly because the number of triggers is much larger than event arguments for all datasets. 

We further look at \texttt{target} and \texttt{fname} prediction scores of the English development set. As shown in \cref{tab:res_detailed}, for \texttt{fname}, our systems tend to over-predict, with consistently lower precision scores; by manually going through our systems' predictions, we find many labeled chunks of \texttt{fname} are actually non-event components. For \texttt{target}, our systems tend to under-predict, with consistently higher precision scores; we also find that our systems would predict a longer span, for instance ``former diplomat'' as opposed to ``diplomat'', which is the gold span, and sometimes our systems confuse \texttt{organizer} and \texttt{participant} with \texttt{target}, by wrongly labelling the corresponding span as \texttt{target}.

\section{Conclusion}
In this paper we have presented the EventGraph system for event extraction and its application to the CASE 2022 shared task on Multilingual Protest Event Detection. EventGraph solves the task as a graph parsing problem hence we experiment with different ways of encoding the event data as general graphs, contrasting a so-called ``labeled-edge'' and ``node-centric'' approach. Our results indicate that the ``node-centric'' approach is beneficial for this task and furthermore that the separation in the graph of nodes belonging to different events in the same sentence proves useful. A more detailed analysis of the development results indicates that our system performs well in trigger identification, however struggles in the identification of \texttt{target} and \texttt{fname} arguments.

\section*{Acknowledgments}
This research was supported by industry partners and the Research Council of Norway with funding to \textit{MediaFutures: Research Centre for Responsible Media Technology and Innovation}, through the Centres for Research-based Innovation scheme, project number 309339.

\bibliography{anthology,custom}
\bibliographystyle{acl_natbib}




\end{document}